# Warped Mixtures for Nonparametric Cluster Shapes

**Tomoharu Iwata**
University of Cambridge
iwata.tomoharu@lab.ntt.co.jp

**David Duvenaud**
University of Cambridge
dkd23@cam.ac.uk

**Zoubin Ghahramani**
University of Cambridge
zoubin@eng.cam.ac.uk

## Abstract

A mixture of Gaussians fit to a single curved or heavy-tailed cluster will report that the data contains many clusters. To produce more appropriate clusterings, we introduce a model which warps a latent mixture of Gaussians to produce nonparametric cluster shapes. The possibly low-dimensional latent mixture model allows us to summarize the properties of the high-dimensional clusters (or density manifolds) describing the data. The number of manifolds, as well as the shape and dimension of each manifold is automatically inferred. We derive a simple inference scheme for this model which analytically integrates out both the mixture parameters and the warping function. We show that our model is effective for density estimation, performs better than infinite Gaussian mixture models at recovering the true number of clusters, and produces interpretable summaries of high-dimensional datasets.

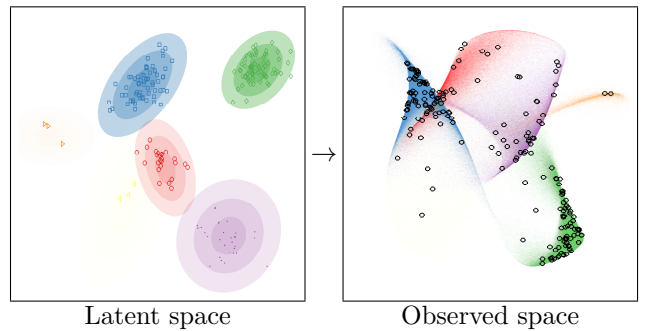

Figure 1: A sample from the iWMM prior. Left: In the latent space, a mixture distribution is sampled from a Dirichlet process mixture of Gaussians. Right: The latent mixture is smoothly warped to produce non-Gaussian manifolds in the observed space.

## 1 Introduction

Probabilistic mixture models are often used for clustering. However, if the mixture components are parametric (e.g. Gaussian), then the clustering obtained can be heavily dependent on how well each actual cluster can be modeled by a Gaussian. For example, a heavy tailed or curved cluster may need many components to model it. Thus, although mixture models are widely used for probabilistic clustering, their assumptions are generally inappropriate if the primary goal is to discover clusters in data. Dirichlet process mixture models can alleviate the problem of an unknown number of clusters, but this does not address the problem that real clusters may not be well matched by any parametric density.

In this paper, we propose a nonparametric Bayesian model that can find nonlinearly separable clusters with complex shapes. The proposed model assumes that each observation has coordinates in a latent space, and is generated by warping the latent coordinates via a nonlinear function from the latent space to the observed space. By this warping, complex shapes in the observed space can be modeled by simpler shapes in the latent space. In the latent space, we assume an infinite Gaussian mixture model [1], which allows us to automatically infer the number of clusters. For the prior on the nonlinear mapping function, we use Gaussian processes [2], which enable us to flexibly infer the nonlinear warping function from the data. We call the proposed model the *infinite warped mixture model* (iWMM). Figure 1 shows a set of manifolds and datapoints sampled from the prior defined by this model.

To our knowledge this is the first probabilistic generative model for clustering with flexible nonparametric component densities. Since the proposed model is generative, it can be used for density estimation as well as clustering. It can also be extended to handle missing data, integrate with other probabilistic models, and use other families of distributions for the latent components.

We derive an inference procedure for the iWMM based on Markov chain Monte Carlo (MCMC). In particular, we sample the cluster assignments using Gibbs sampling, sample the latent coordinates using hybrid Monte Carlo, and analytically integrate out both the mixture parameters (weights, means and covariance matrices), and the nonlinear warping function.

## 2 Gaussian Process Latent Variable Model

In this section, we give a brief introduction to the Gaussian process latent variable model (GPLVM) [3], which can be viewed as a special case of the iWMM. The GPLVM is a probabilistic model of nonlinear manifolds. While not typically thought of as a density model, the GPLVM does in fact define a posterior density over observations [4]. It does this by smoothly warping a single, isotropic Gaussian density in the latent space into a more complicated distribution in the observed space.

Suppose that we have a set of observations $\mathbf{Y} = (\mathbf{y}_1, \cdots, \mathbf{y}_N)^\top$, where $\mathbf{y}_n \in \mathbb{R}^D$, and they are associated with a set of latent coordinates $\mathbf{X} = (\mathbf{x}_1, \cdots, \mathbf{x}_N)^\top$, where $\mathbf{x}_n \in \mathbb{R}^Q$. The GPLVM assumes that observations are generated by mapping the latent coordinates through a set of smooth functions, over which Gaussian process priors are placed. Under the GPLVM, the probability of observations given the latent coordinates, integrating out the mapping functions, is

$$p(\mathbf{Y}|\mathbf{X}, \boldsymbol{\theta}) = (2\pi)^{-\frac{DN}{2}} |\mathbf{K}|^{-\frac{D}{2}} \exp\left(-\frac{1}{2}\mathrm{tr}(\mathbf{Y}^\top \mathbf{K}^{-1} \mathbf{Y})\right), \quad (1)$$

where $\mathbf{K}$ is the $N \times N$ covariance matrix defined by the kernel function $k(\mathbf{x}_n, \mathbf{x}_m)$, and $\boldsymbol{\theta}$ is the kernel hyperparameter vector. In this paper, we use an RBF kernel with an additive noise term:

$$k(\mathbf{x}_n, \mathbf{x}_m) = \alpha \exp\left(-\frac{1}{2\ell^2}(\mathbf{x}_n - \mathbf{x}_m)^\top(\mathbf{x}_n - \mathbf{x}_m)\right) + \delta_{nm}\beta^{-1}. \quad (2)$$

This likelihood is simply the product of $D$ independent Gaussian process likelihoods, one for each output dimension.

Typically, the GPLVM is used for dimensionality reduction or visualization, and the latent coordinates are determined by maximizing the posterior probability of the latent coordinates, while integrating out the warping function. In that setting, the Gaussian prior density on $\mathbf{x}$ is essentially a regularizer which keeps the latent coordinates from spreading arbitrarily far apart. In contrast, we instead integrate out the latent coordinates as well as the warping function, and place a more flexible parameterization on $p(\mathbf{x})$ than a single isotropic Gaussian.

Just as the GPLVM can be viewed as a manifold learning algorithm, the iWMM can be viewed as learning a set of manifolds, one for each cluster.

## 3 Infinite Warped Mixture Model

In this section, we define in detail the infinite warped mixture model (iWMM). In the same way as the GPLVM, the iWMM assumes a set of latent coordinates and a smooth, nonlinear mapping from the latent space to the observed space. In addition, the iWMM assumes that the latent coordinates are generated from a Dirichlet process mixture model. In particular, we use the following infinite Gaussian mixture model,

$$p(\mathbf{x}|\{\lambda_c, \boldsymbol{\mu}_c, \mathbf{R}_c\}) = \sum_{c=1}^{\infty} \lambda_c \mathcal{N}(\mathbf{x}|\boldsymbol{\mu}_c, \mathbf{R}_c^{-1}), \quad (3)$$

where $\lambda_c$, $\boldsymbol{\mu}_c$ and $\mathbf{R}_c$ is the mixture weight, mean, and precision matrix of the $c^{\mathrm{th}}$ mixture component. We place Gaussian-Wishart priors on the Gaussian parameters $\{\boldsymbol{\mu}_c, \mathbf{R}_c\}$,

$$p(\boldsymbol{\mu}_c, \mathbf{R}_c) = \mathcal{N}(\boldsymbol{\mu}_c|\mathbf{u}, (r\mathbf{R}_c)^{-1})\mathcal{W}(\mathbf{R}_c|\mathbf{S}^{-1}, \nu), \quad (4)$$

where $\mathbf{u}$ is the mean of $\boldsymbol{\mu}_c$, $r$ is the relative precision of $\boldsymbol{\mu}_c$, $\mathbf{S}^{-1}$ is the scale matrix for $\mathbf{R}_c$, and $\nu$ is the number of degrees of freedom for $\mathbf{R}_c$. The Wishart distribution is defined as follows:

$$\mathcal{W}(\mathbf{R}|\mathbf{S}^{-1}, \nu) = \frac{1}{G}|\mathbf{R}|^{\frac{\nu-Q-1}{2}} \exp\left(-\frac{1}{2}\mathrm{tr}(\mathbf{S}\mathbf{R})\right), \quad (5)$$

where $G$ is the normalizing constant. Because we use conjugate Gaussian-Wishart priors for the parameters of the Gaussian mixture components, we can analytically integrate out those parameters, given the assignments of points to components. Let $z_n$ be the latent assignment of the $n^{\mathrm{th}}$ point. The probability of latent coordinates $\mathbf{X}$ given latent assignments $\mathbf{Z} = (z_1, \cdots, z_N)$ is obtained by integrating out the Gaussian parameters $\{\boldsymbol{\mu}_c, \mathbf{R}_c\}$ as follows:

$$p(\mathbf{X}|\mathbf{Z}, \mathbf{S}, \nu, r) = \prod_{c=1}^{\infty} \pi^{-\frac{N_c Q}{2}} \frac{r^{Q/2}|\mathbf{S}|^{\nu/2}}{r_c^{Q/2}|\mathbf{S}_c|^{\nu_c/2}} \\ \times \prod_{q=1}^{Q} \frac{\Gamma(\frac{\nu_c+1-q}{2})}{\Gamma(\frac{\nu+1-q}{2})}, \quad (6)$$

where $N_c$ is the number of data points assigned to the $c^{\mathrm{th}}$ component, $\Gamma(\cdot)$ is Gamma function, and

$$r_c = r + N_c, \qquad \nu_c = \nu + N_c,$$

$$\mathbf{u}_c = \frac{r\mathbf{u} + \sum_{n:z_n=c}\mathbf{x}_n}{r + N_c},$$

$$\mathbf{S}_c = \mathbf{S} + \sum_{n:z_n=c}\mathbf{x}_n\mathbf{x}_n^\top + r\mathbf{u}\mathbf{u}^\top - r_c\mathbf{u}_c\mathbf{u}_c^\top, \quad (7)$$

are the posterior Gaussian-Wishart parameters of the $c^{\text{th}}$ component. We use a Dirichlet process with concentration parameter $\eta$ for infinite mixture modeling [5] in the latent space. Then, the probability of $\mathbf{Z}$ is given as follows:

$$p(\mathbf{Z}|\eta) = \frac{\eta^C \prod_{c=1}^{C}(N_c - 1)!}{\eta(\eta+1)\cdots(\eta+N-1)}, \quad (8)$$

where $C$ is the number of components for which $N_c > 0$. The joint distribution is given by

$$\begin{aligned}&p(\mathbf{Y},\mathbf{X},\mathbf{Z}|\boldsymbol{\theta},\boldsymbol{S},\nu,\mathbf{u},r,\eta)\\&= p(\mathbf{Y}|\mathbf{X},\boldsymbol{\theta})p(\mathbf{X}|\mathbf{Z},\boldsymbol{S},\nu,\mathbf{u},r)p(\mathbf{Z}|\eta),\end{aligned} \quad (9)$$

where factors in the right hand side can be calculated by (1), (6) and (8), respectively.

In summary, the infinite warped mixture model generates observations $\mathbf{Y}$ according to the following generative process:

1. Draw mixture weights $\boldsymbol{\lambda} \sim \text{GEM}(\eta)$
2. For each component $c = 1, \cdots, \infty$
   (a) Draw precision $\mathbf{R}_c \sim \mathcal{W}(\mathbf{S}^{-1}, \nu)$
   (b) Draw mean $\boldsymbol{\mu}_c \sim \mathcal{N}(\mathbf{u}, (r\mathbf{R}_c)^{-1})$
3. For each observed dimension $d = 1, \cdots, D$
   (a) Draw function $f_d(\mathbf{x}) \sim \text{GP}(m(\mathbf{x}), k(\mathbf{x}, \mathbf{x}'))$
4. For each observation $n = 1, \cdots, N$
   (a) Draw latent assignment $z_n \sim \text{Mult}(\boldsymbol{\lambda})$
   (b) Draw latent coordinates $\mathbf{x}_n \sim \mathcal{N}(\boldsymbol{\mu}_{z_n}, \mathbf{R}_{z_n}^{-1})$
   (c) For each observed dimension $d = 1, \cdots, D$
      i. Draw feature $y_{nd} \sim \mathcal{N}(f_d(\mathbf{x}_n), \beta^{-1})$

Here, $\text{GEM}(\eta)$ is the stick-breaking process [6] that generates mixture weights for a Dirichlet process with parameter $\eta$, $\text{Mult}(\boldsymbol{\lambda})$ represents a multinomial distribution with parameter $\boldsymbol{\lambda}$, $m(\mathbf{x})$ is the mean function of the Gaussian process, and $\mathbf{x}, \mathbf{x}' \in \mathbb{R}^Q$. Figure 2 shows the graphical model representation of the proposed model. Here, we assume a Gaussian for the mixture component, although we could in principle use other distributions such as Student's t-distribution or the Laplace distribution.

The iWMM can be seen as a generalization of either the GPLVM or the infinite Gaussian mixture model

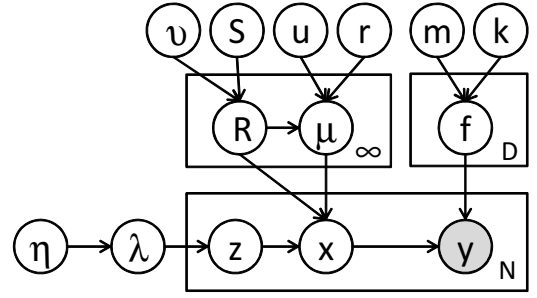

Figure 2: A graphical model representation of the infinite warped mixture model, where the shaded and unshaded nodes indicate observed and latent variables, respectively, and plates indicate repetition.

(iGMM). To be precise, the iWMM with a single fixed spherical Gaussian density on the latent coordinates corresponds to the GPLVM, while the iWMM with fixed direct mapping function $f_d(\mathbf{x}) = x_d$ and $Q = D$ corresponds to the iGMM.

The iWMM offers attractive properties that do not exist in other probabilistic models; principally, the ability to model clusters with nonparametric densities, and to infer a separate dimension for manifold.

## 4 Inference

We infer the posterior distribution of the latent coordinates $\mathbf{X}$ and cluster assignments $\mathbf{Z}$ using Markov chain Monte Carlo (MCMC). In particular, we alternate collapsed Gibbs sampling of $\mathbf{Z}$, and hybrid Monte Carlo sampling of $\mathbf{X}$. Given $\mathbf{X}$, we can efficiently sample $\mathbf{Z}$ using collapsed Gibbs sampling, integrating out the mixture parameters. Given $\mathbf{Z}$, we can calculate the gradient of the unnormalized posterior distribution of $\mathbf{X}$, integrating over warping functions. This gradient allows us to sample $\mathbf{X}$ using hybrid Monte Carlo.

First, we explain collapsed Gibbs sampling for $\mathbf{Z}$. Given a sample of $\mathbf{X}$, $p(\mathbf{Z}|\mathbf{X},\mathbf{S},\nu,\mathbf{u},r,\eta)$ does not depend on $\mathbf{Y}$. This lets resample cluster assignments, integrating out the iGMM likelihood in close form. Given the current state of all but one latent component $z_n$, a new value for $z_n$ is sampled from the following probability:

$$\begin{aligned}&p(z_n = c|\mathbf{X},\mathbf{Z}_{\setminus n},\boldsymbol{S},\nu,\mathbf{u},r,\eta)\\&\propto \begin{cases} N_{c\setminus n} \cdot p(\mathbf{x}_n|\mathbf{X}_{c\setminus n},\boldsymbol{S},\nu,\mathbf{u},r) & \text{existing components} \\ \eta \cdot p(\mathbf{x}_n|\boldsymbol{S},\nu,\mathbf{u},r) & \text{a new component} \end{cases}\end{aligned}$$
$$(10)$$

where $\mathbf{X}_c = \{\mathbf{x}_n | z_n = c\}$ is the set of latent coordinates assigned to the $c^{\text{th}}$ component, and $\setminus n$ represents the value or set when excluding the $n^{\text{th}}$ data point. We can analytically calculate $p(\mathbf{x}_n|\mathbf{X}_{c\backslash n}, \boldsymbol{S}, \nu, \mathbf{u}, r)$ as follows:

$$p(\mathbf{x}_n|\mathbf{X}_{c\backslash n}, \boldsymbol{S}, \nu, \mathbf{u}, r)$$
$$= \pi^{-\frac{N_{c\backslash n}Q}{2}} \frac{r_{c\backslash n}^{Q/2}|\mathbf{S}_{c\backslash n}|^{\nu_{c\backslash n}/2}}{r_{c\backslash n}^{\prime Q/2}|\mathbf{S}_{c\backslash n}'|^{\nu_{c\backslash n}'/2}} \prod_{d=1}^{Q} \frac{\Gamma(\frac{\nu_{c\backslash n}'+1-d}{2})}{\Gamma(\frac{\nu_{c\backslash n}+1-d}{2})}, \quad (11)$$

where $r_c'$, $\nu_c'$, $\mathbf{u}_c'$ and $\mathbf{S}_c'$ represent the posterior Gaussian-Wishart parameters of the $c^{\text{th}}$ component when the $n^{\text{th}}$ data point is assigned to the $c^{\text{th}}$ component. We can efficiently calculate the determinant by using the rank one Cholesky update. In the same way, we can analytically calculate the likelihood for a new component $p(\mathbf{x}_n|\boldsymbol{S}, \nu, \mathbf{u}, r)$.

Hybrid Monte Carlo (HMC) sampling of $\mathbf{X}$ from posterior $p(\mathbf{X}|\mathbf{Z}, \mathbf{Y}, \boldsymbol{\theta}, \boldsymbol{S}, \nu, \mathbf{u}, r)$ requires computing the gradient of the log of the unnormalized posterior $\log p(\mathbf{Y}|\mathbf{X}, \boldsymbol{\theta}) + \log p(\mathbf{X}|\mathbf{Z}, \boldsymbol{S}, \nu, \mathbf{u}, r)$. The first term of the gradient can be calculated by

$$\frac{\partial \log p(\mathbf{Y}|\mathbf{X}, \boldsymbol{\theta})}{\partial \mathbf{K}} = -\frac{1}{2}D\mathbf{K}^{-1} + \frac{1}{2}\mathbf{K}^{-1}\mathbf{Y}\mathbf{Y}^T\mathbf{K}^{-1}, \quad (12)$$

and

$$\frac{\partial k(\mathbf{x}_n, \mathbf{x}_m)}{\partial \mathbf{x}_n}$$
$$= -\frac{\alpha}{\ell^2} \exp\left(-\frac{1}{2\ell^2}(\mathbf{x}_n - \mathbf{x}_m)^\top(\mathbf{x}_n - \mathbf{x}_m)\right)(\mathbf{x}_n - \mathbf{x}_m), \quad (13)$$

using the chain rule. The second term can be calculated as follows:

$$\frac{\partial \log p(\mathbf{X}|\mathbf{Z}, \boldsymbol{S}, \nu, \mathbf{u}, r)}{\partial \mathbf{x}_n} = -\nu_{z_n} \boldsymbol{S}_{z_n}^{-1}(\mathbf{x}_n - \mathbf{u}_{z_n}). \quad (14)$$

We also infer kernel hyperparameters $\boldsymbol{\theta} = \{\alpha, \beta, \ell\}$ via HMC, using the gradient of the log unnormalized posterior with respect to the kernel hyperparameters. The complexity of each iteration of HMC is dominated by the $\mathcal{O}(N^3)$ computation of $\mathbf{K}^{-1}$ [1].

In summary, we obtain samples from the posterior $p(\mathbf{X}, \mathbf{Z}|\mathbf{Y}, \boldsymbol{\theta}, \boldsymbol{S}, \nu, \mathbf{u}, r, \eta)$ by iterating the following procedures:

1. For each observation $n = 1, \cdots, N$, sample the component assignment $z_n$ by collapsed Gibbs sampling (10).

2. Sample latent coordinates $\mathbf{X}$ and kernel parameters $\boldsymbol{\theta}$ using hybrid Monte Carlo.

---
[1] This complexity could be improved by making use of an inducing point approximation such as [7, 8]

## 4.1 Posterior Predictive Density

In the GPLVM, the predictive density of at test point $y^\star$ is usually computed by finding the point $x^\star$ which is most likely to be mapped to $y^\star$, then using the density of $p(x^\star)$ and the Jacobian of the warping at that point to approximately compute the density at $y^\star$. When inference is done by simply optimizing the location of the latent points, this estimation method simply requires solving a single optimization for each $y^\star$.

For our model, we use approximate integration to estimate $p(y^\star)$. This is done for two reasons: First, multiple latent points (possibly from different clusters) can map to the same observed point, meaning the standard method can underestimate $p(y^\star)$. Second, because we do not optimize the latent coordinates but rather sample them, we would need to perform optimizations for each $p(y^\star)$ separately for each sample. Our method gives estimates for all $p(y^\star)$ at once, but may not be accurate in very high dimensions.

The posterior density in the observed space given the training data is simply:

$$p(\mathbf{y}^\star|\mathbf{Y})$$
$$= \iint p(\mathbf{y}^\star, \mathbf{x}^\star, \mathbf{X}|\mathbf{Y})d\mathbf{x}^\star d\mathbf{X}$$
$$= \iint p(\mathbf{y}^\star|\mathbf{x}^\star, \mathbf{X}, \mathbf{Y})p(\mathbf{x}^\star|\mathbf{X}, \mathbf{Y})p(\mathbf{X}|\mathbf{Y})d\mathbf{x}^\star d\mathbf{X}. \quad (15)$$

We approximate $p(\mathbf{X}|\mathbf{Y})$ using the samples from the Gibbs and hybrid Monte Carlo samplers. We approximate $p(\mathbf{x}^\star|\mathbf{X}, \mathbf{Y})$ by sampling points from the latent mixture and warping them, using the following procedure:

1. Draw latent assignment
   $z^\star \sim \text{Mult}(\frac{N_1}{N+\eta}, \cdots, \frac{N_C}{N+\eta}, \frac{\eta}{N+\eta})$

2. Draw precision matrix
   $\mathbf{R}^\star \sim \mathcal{W}(\mathbf{S}_{z^\star}^{-1}, \nu_{z^\star})$

3. Draw mean
   $\boldsymbol{\mu}^\star \sim \mathcal{N}(\mathbf{u}_{z^\star}, (r_{z^\star}\mathbf{R}^\star)^{-1})$

4. Draw latent coordinates
   $\mathbf{x}^\star \sim \mathcal{N}(\boldsymbol{\mu}^\star, \mathbf{R}^{\star -1})$

When a new component $C + 1$ is assigned to $z^\star$, the prior Gaussian-Wishart distribution is used for sampling in steps 2 and 3. The first factor of (15) can be calculated by

$$p(\mathbf{y}^\star|\mathbf{x}^\star, \mathbf{X}, \mathbf{Y})$$
$$= \mathcal{N}(\mathbf{k}^{\star\top}\mathbf{K}^{-1}\mathbf{Y}, k(\mathbf{x}^\star, \mathbf{x}^\star) - \mathbf{k}^{\star\top}\mathbf{K}^{-1}\mathbf{k}^\star), \quad (16)$$

where $\mathbf{k}^\star = (k(\mathbf{x}^\star, \mathbf{x}_1), \cdots, k(\mathbf{x}^\star, \mathbf{x}_N))^\top$. Each step of this procedure is exact. Since the observations $\mathbf{y}^\star$ are conditionally normally distributed, each one adds a smooth local contribution to the empirical Monte Carlo estimate of the posterior density, as opposed to a point mass. This procedure was used to generate the plots of posterior density in Figures 1, 4, and 6.

## 5 Related work

**Latent Variable Models**  The GPLVM is effective as a nonlinear latent variable model in a wide variety of applications [3, 9, 10]. The latent positions $\mathbf{X}$ in the GPLVM are typically obtained by maximum a posteriori estimation or variational Bayesian inference [11], placing a single fixed spherical Gaussian prior on $\mathbf{x}$. A prior which penalizes a high-dimensional latent space is introduced by [12], in which the latent variables and their intrinsic dimensionality are simultaneously optimized. The iWMM can also infer the intrinsic dimensionality of each manifolds: inferring the Gaussian covariance for each latent cluster allows the variance along irrelevant dimensions to become small. Because each latent cluster has a different set of parameters, the effective dimension of each cluster can vary, allowing manifolds of different dimension in the observed space. This ability is demonstrated in Figure 4 (c).

The iWMM can also be viewed as a generalization of the mixture of probabilistic principle component analyzers [13], or mixture of factor analyzers [14], where the linear mapping of the mixtures is generalized to a nonlinear mapping by Gaussian processes, and number of components is infinite.

**Clustering Methods**  There exist non-probabilistic clustering methods which can find clusters with complex shapes, such as spectral clustering [15] and nonlinear manifold clustering [16, 17]. Spectral clustering finds clusters by first forming a similarity graph, then finding a low-dimensional latent representation using the graph, and finally, clustering the latent coordinates via k-means. The performance of spectral clustering depends on parameters which are usually set manually, such as the number of clusters, the number of neighbors, and the variance parameter used for constructing the similarity graph. In contrast, the iWMM infers such parameters automatically. One of the main advantages of the iWMM over these methods is that there is no need to construct a similarity graph.

The kernel Gaussian mixture model [18] can also find non-Gaussian shaped clusters. This model estimates a GMM in the implicit high-dimensional feature space defined by the kernel mapping of the observed space. However, the kernel GMM uses a fixed nonlinear map-ping function, with no guarantee that the latent points will be well-modeled by a GMM. In contrast, the iWMM infers the mapping function such that the latent co-ordinates will be well-modeled by a GMM.

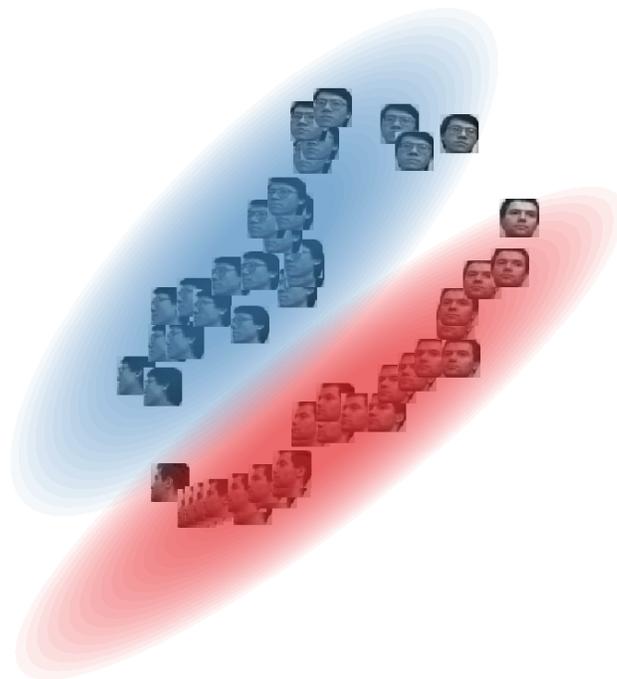

Figure 3: A sample from the 2-dimensional latent space when modeling a series of 32x32 face images. Our model correctly discovers that the data consists of two separate manifolds, both approximately one-dimensional, which share the same head-turning structure.

For one-dimensional data, [19] introduce a nonparametric model of *unimodal* clusters, where each cluster's density function decreases away from its mode.

## 6 Experimental results

**Clustering Faces**  We first examined our model's ability to model images without pre-processing. We constructed a dataset consisting of 50 greyscale 32x32 pixel images of two individuals from the UMIST faces dataset [20]. Both series of images capture a person turning his head to the right. Figure 3 shows a sample from the posterior over the latent coordinates and density model. The model has recovered three relevant, interpretable features of the dataset. First, that there are two distinct faces. Second, that each set of images lies approximately along a smooth one-dimensional manifold. Third, that the two manifolds share roughly the same structure: the front-facing images of both individuals lie close to one another, as do the side-facing images.

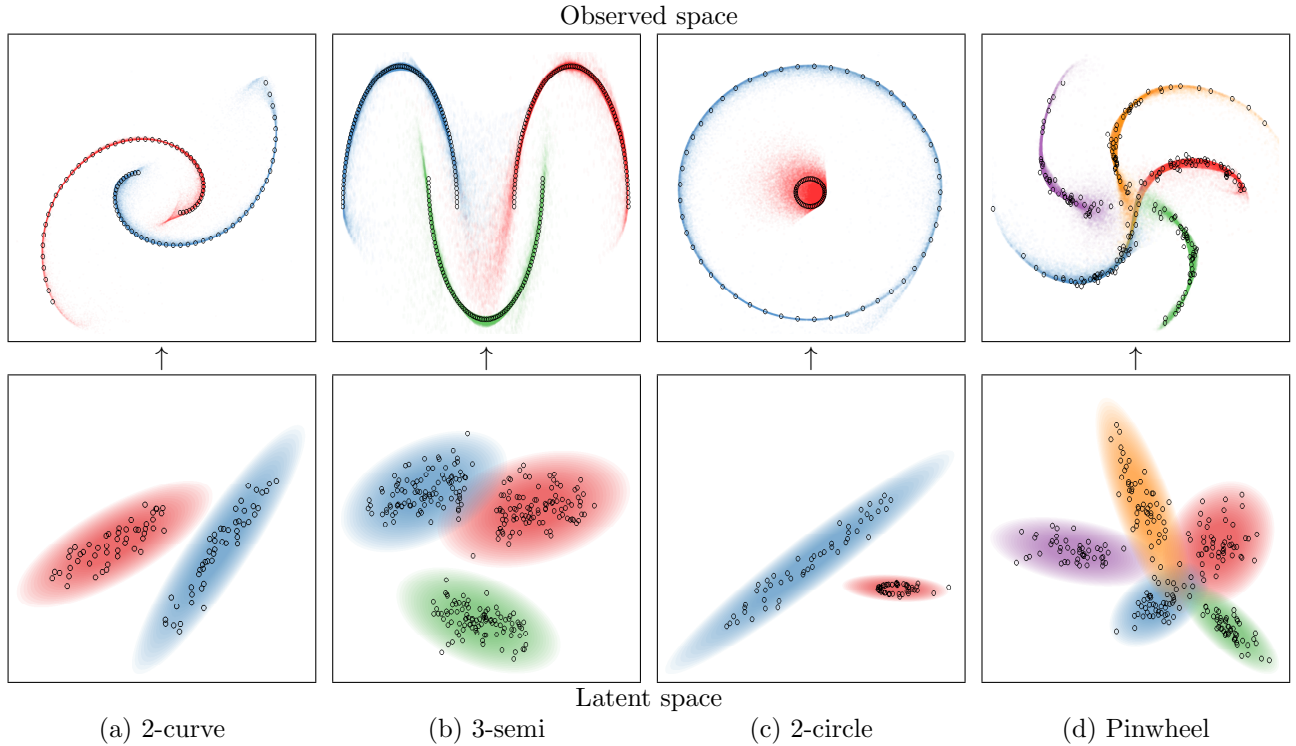

Figure 4: Top row: The observed, unlabeled data points, and the clusters inferred by the iWMM. Bottom row: Latent coordinates and Gaussian components, shown for a single sample from the posterior. Each point in the latent space corresponds to a point in the observed space. This figure is best viewed in color.

### 6.1 Synthetic Datasets

Next, we demonstrate the proposed model on the four synthetic datasets shown in Figure 4. None of these four datasets can be appropriately clustered by Gaussian mixture models (GMM). For example, consider the 2-curve data shown in Figure 4 (a), where 100 data points lie in one of two curved lines in a two-dimensional observed space. A GMM with two components cannot separate the two curved lines, while a GMM with many components could separate the two lines only by breaking each line into many clusters. In contrast, in the iWMM, the two non-Gaussian-shaped clusters in the observed space were represented by two Gaussian-shaped clusters in the latent space, as shown at the bottom row of Figure 4 (a). The iWMM separated the two curved lines by nonlinearly warping two Gaussians from the latent space to the observed space.

Figure 4 (c) shows an interesting manifold learning challenge: a dataset consisting of two circles. The outer circle is modeled in the latent space by a Gaussian with effectively one degree of freedom. This linear topology fits the outer circle in the observed space by bending the two ends until they overlap. In contrast, the sampler fails to discover the 1D topology of the inner circle, modeling it with a 2D manifold instead. This example demonstrates that each cluster manifold in the iWMM can have a different effective dimension.

### 6.2 Mixing

An interesting side-effect of learning the number of latent clusters is that this added flexibility can help the sampler escape local minima, helping the sampler to mix properly. Figure 5 shows the samples of the latent coordinates and clusters of the iWMM over time, when modeling the 2-curve data. 5(a) shows the latent coordinates initialized at the observed coordinates, starting with one latent component. At the 500th iteration 5(b), each curved line is modeled by two components. At the 1800th iteration 5(c), the left curved line is modeled by a single component. At the 3000th iteration 5(d), the right curved line is also modeled by a single component, and the dataset is appropriately clustered. This configuration was relatively stable, and a similar state was found at the 5000th iteration.

### 6.3 Density Estimation

Figure 6 (a) shows the posterior density in the observed space inferred by the iWMM on the 2-curve

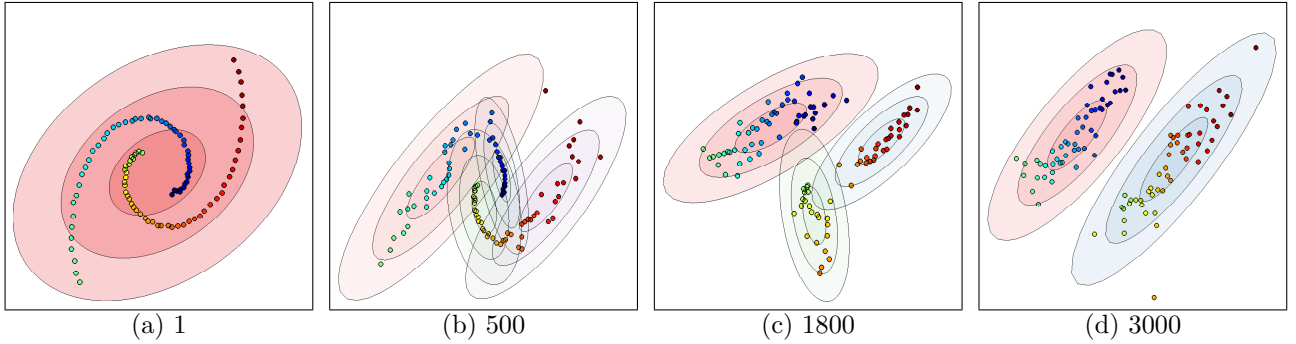

(a) 1     (b) 500     (c) 1800     (d) 3000

Figure 5: The inferred infinite GMMs over iterations in the two-dimensional latent space with the iWMM using the 2-curve data. Labels indicate the number of iterations of the sampler, and the color of each point represents its ordering in the observed coordinates.

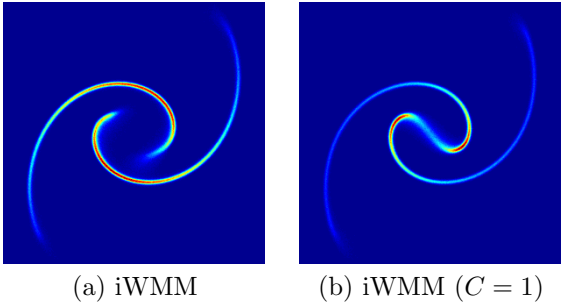

(a) iWMM     (b) iWMM ($C = 1$)

Figure 6: The posterior density in the observed space with the 2-curve data inferred by the iWMM (a), and that inferred by the iWMM with one component (b).

data, computed using 1000 samples from the Markov chain. The two separate manifolds of high density implied by the two curved lines was recovered by the iWMM. Note also that the density along the manifold varies with the density of data shown in Figure 4 (a). This result can be compared to a special case of our model, which uses only a single Gaussian to model the latent coordinates instead of an infinite GMM. Figure 6 (b) shows that the result of the iWMM with $C = 1$, where posterior is forced to place significant density connecting the two clusters. Figure 6 (b) shows that the single-cluster variant of the iWMM posterior is forced to place significant density connecting the two clusters.

### 6.4 Visualization

Next, we briefly investigate the potential of the iWMM for visualization. Figure 7 (a) shows the latent coordinates obtained by averaging over 1000 samples from the posterior of the iWMM. Because rotating the latent coordinates does not change their probability, averaging may not be an adequate way to summarize the posterior. However, we show this result in order to show the characteristics of latent coordinates obtained by the iWMM. The estimated latent coordinates are clearly separated, and they form two straight lines. This result indicates that in some cases, the iWMM can recover the topology of the data before it has been warped into a manifold. For comparison, Figure 7 (b) shows the latent coordinates estimated by the iWMM when forced to use a single cluster: the latent coordinates lie in two sections of a single straight line. Figures 7 (c) and (d) show the latent coordinates estimated by the GPLVM when optimizing or integrating out the latent coordinates, respectively. Recall that the iWMM ($C = 1$) is a more flexible model than the GPLVM, since the GPLVM enforces a spherical covariance in the latent space. These methods did not unfold the two curved lines, since the effective dimension of their latent representation is fixed beforehand. In contrast, the iWMM effectively formed a low-dimensional representation in the latent space.

Regardless of the dimension of the latent space, the iWMM will tend to model each cluster with as low-dimensional a Gaussian as possible. This is because, if the data in a cluster can be made to lie in a low-dimensional plane, a narrowly-shaped Gaussian will assign the latent coordinates much higher likelihood than a spherical Gaussian.

### 6.5 Clustering Performance

We more formally evaluated the density estimation and clustering performance of the proposed model using four real datasets: iris, glass, wine and vowel, obtained from LIBSVM multi-class datasets [21], in addition to the four synthetic datasets shown above: 2-curve, 3-semi, 2-circle and Pinwheel [22]. The statistics of these datasets are summarized in Table 1. In each experiment, we show the results of 20-fold cross-validation. Results in bold are not significantly different from the best performing method in each column

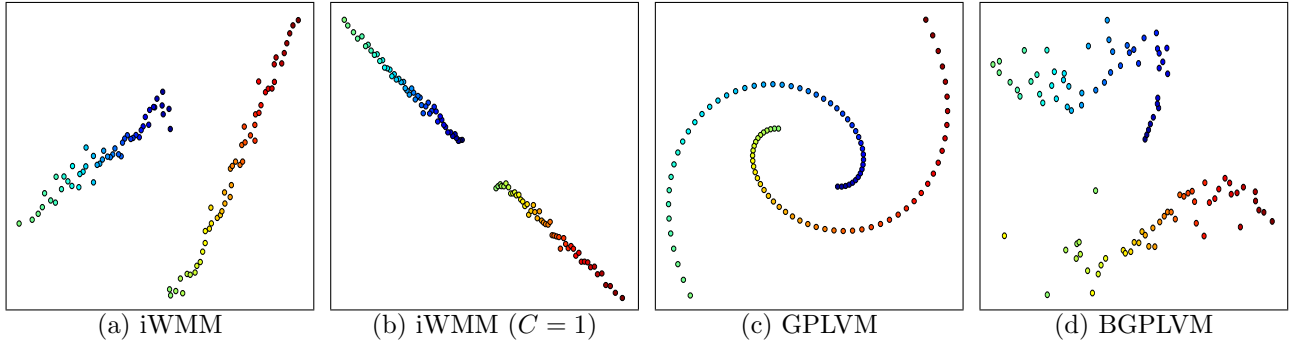

(a) iWMM  (b) iWMM ($C = 1$)  (c) GPLVM  (d) BGPLVM

Figure 7: The estimated latent coordinates of the 2-curve data by (a) iWMM, (b) iWMM ($C = 1$), (c) GPLVM, and (d) Bayesian GPLVM.

Table 1: The statistics of datasets used for evaluation.

|   | 2-curve | 3-semi | 2-circle | Pinwheel | Iris | Glass | Wine | Vowel |
|---|---|---|---|---|---|---|---|---|
| number of samples: $N$ | 100 | 300 | 100 | 250 | 150 | 214 | 178 | 528 |
| observed dimensionality: $D$ | 2 | 2 | 2 | 2 | 4 | 9 | 13 | 10 |
| number of clusters: $C$ | 2 | 3 | 2 | 5 | 3 | 7 | 3 | 11 |

Table 2: Average Rand index for evaluating clustering performance.

|   | 2-curve | 3-semi | 2-circle | Pinwheel | Iris | Glass | Wine | Vowel |
|---|---|---|---|---|---|---|---|---|
| iGMM | 0.52 | 0.79 | 0.83 | 0.81 | 0.78 | 0.60 | 0.72 | **0.76** |
| iWMM(Q=2) | **0.86** | **0.99** | **0.89** | **0.94** | **0.81** | **0.65** | 0.65 | 0.50 |
| iWMM(Q=D) | **0.86** | **0.99** | **0.89** | **0.94** | 0.77 | 0.62 | **0.77** | **0.76** |

according to a paired t-test.

Table 2 compares the clustering performance of the iWMM with the iGMM, quantified by the Rand index [23], which measures the correspondence between inferred clusters and true clusters. The iGMM is another probabilistic generative model commonly used for clustering, which can be seen as a special case of the iWMM in which the Gaussian clusters are not warped. These experiments demonstrate the extent to which nonparametric cluster shapes allow a mixture model to recover more meaningful clusters.

Table 3 lists average test log likelihood, comparing the proposed models with kernel density estimation (KDE), and the infinite Gaussian mixture model (iGMM). In KDE, the kernel width is estimated by maximizing the leave-one-out log densities. Since the manifold on which the observed data lies can be at most $D$-dimensional, we set the latent dimension $Q$ equal to the observed dimension $D$ in iWMMs. We also include the $Q = 2$ case in an attempt to characterize how much modeling power is lost by forcing the latent representation to be visualizable. The proposed models achieved high test log likelihoods compared with the KDE and iGMM.

### 6.6 Source code

Code to reproduce all the above experiments is available at github.com/duvenaud/warped-mixtures.

## 7 Future work

The Dirichlet process mixture of Gaussians in the latent space of our model could easily be replaced by a more sophisticated density model, such as a hierarchical Dirichlet process [24], or a Dirichlet diffusion tree [25]. Another straightforward extension of our model would be making inference more scalable by using sparse Gaussian processes [7, 8] or more advanced hybrid Monte Carlo methods [26]. An interesting but more complex extension of the iWMM would be a semi-supervised version of the model. The iWMM could allow label propagation along regions of high density in the latent space, even if those regions were stretched along low-dimensional manifolds in the observed space. Another natural extension would be to allow a separate warping for each cluster, which would also improve inference speed.

Table 3: Average test log likelihood for evaluating density estimation performance.

|            | 2-curve | 3-semi | 2-circle | Pinwheel | Iris   | Glass | Wine   | Vowel |
|------------|---------|--------|----------|----------|--------|-------|--------|-------|
| KDE        | −2.47   | −0.38  | −1.92    | −1.47    | **−1.87** | 1.26  | −2.73  | **6.06** |
| iGMM       | −3.28   | −2.26  | −2.21    | −2.12    | −1.91  | 3.00  | **−1.87** | −0.67 |
| iWMM(Q=2)  | **−0.90** | **−0.18** | **−1.02** | **−0.79** | −1.88  | **5.76** | −1.96  | 5.91  |
| iWMM(Q=D)  | **−0.90** | **−0.18** | **−1.02** | **−0.79** | −1.71  | 5.70  | −3.14  | −0.35 |

## 8 Conclusion

In this paper, we introduced a simple generative model of non-Gaussian density manifolds which can infer nonlinearly separable clusters, low-dimensional representations of varying dimension per cluster, and density estimates which smoothly follow data contours. We then introduced an efficient sampler for this model which integrates out both the cluster parameters and the warping function exactly. We further demonstrated that allowing non-parametric cluster shapes improves clustering performance over the Dirichlet process Mixture of Gaussians.

Many methods have been proposed which can perform some combination of clustering, manifold learning, density estimation and visualization. We demonstrated that a simple but flexible probabilistic generative model can perform well at all these tasks.

**Acknowledgements**

The authors would like to thank Dominique Perrault-Joncas, Carl Edward Rasmussen, and Ryan Prescott Adams for helpful discussions.